\documentclass[conference]{IEEEtran}
\IEEEoverridecommandlockouts

\usepackage{cite}
\usepackage{amsmath,amssymb,amsfonts}
\usepackage{algorithmic}
\usepackage{graphicx}
\usepackage{textcomp}
\usepackage{xcolor}
\def\BibTeX{{\rm B\kern-.05em{\sc i\kern-.025em b}\kern-.08em
    T\kern-.1667em\lower.7ex\hbox{E}\kern-.125emX}}
\begin{document}

\title{Exploring Kolmogorov-Arnold networks 
for realistic image sharpness assessment}

\author{
	\IEEEauthorblockN{Shaode Yu, Ze Chen, Zhimu Yang, Jiacheng Gu, Bizu Feng}
	\IEEEauthorblockA{\textit{School of Information and Communication Engineering} \\
		\textit{Communication University of China}\\
		Beijing, China \\
		\{yushaodecuc, chenze, 2021211123030\}@cuc.edu.cn}
	\and
	\IEEEauthorblockN{Qiurui Sun*}
	\IEEEauthorblockA{\textit{Center of Information \& Network Technology} \\
		\textit{Beijing Normal University}\\
		Beijing, China \\
		qiuruisun@bnu.edu.cn}
}

\maketitle

\begin{abstract}
Score prediction is crucial in evaluating realistic image sharpness based on collected informative features. Recently, Kolmogorov-Arnold networks (KANs) have been developed and witnessed remarkable success in data fitting. This study introduces the Taylor series-based KAN (TaylorKAN). Then, different KANs are explored in four realistic image databases (BID2011, CID2013, CLIVE, and KonIQ-10k) to predict the scores by using 15 mid-level features and 2048 high-level features. Compared to support vector regression, results show that KANs are generally competitive or superior, and TaylorKAN is the best one when mid-level features are used. This is the first study to investigate KANs on image quality assessment that sheds some light on how to select and further improve KANs in related tasks. 
\end{abstract}


\section{Introduction}
\label{sec:intro}
%
Blind image sharpness/blurriness assessment (BISA) is crucial in media quality assurance. 
It enables real-time processing without reference images and supports dynamic adjustments to image sharpness and visual fidelity in video streaming, 
thereby enhancing the quality of the user experience. \cite{zhai2020perceptual}. 
Specifically, BISA can be used to guide the restoration process by offering a metric to optimize algorithms for image sharpening \cite{li2020deep}.

%
Score prediction is an indispensable step when informative features have been prepared for image sharpness representation. 
Support vector regression (SVR) and multi-layer perceptron (MLP) are preferred. 
Li $et$ $al$ craft multi-scale sharpness-aware features, and SVR performs score rating \cite{li2016no}. 
Yu $et$ $al$ design a convolutional neural network (CNN) for blurriness estimation \cite{yu2016cnn}, and SVR and MLP are used to improve the prediction performance \cite{yu2017shallow}. 
Liu $et$ $al$ design orientation-aware features for SVR-based score prediction \cite{liu2020blind}. 
Chen $et$ $al$ weight the local binary pattern features in spatial domain and entropy and gradient features in spectral domain, and SVR predicts the quality scores from perception features \cite{chen2023no}. 
Yu $et$ $al$ construct mid-level features, and MLP and SVR are evaluated \cite{yu2023hybrid}.

For score prediction, a CNN model can be generally treated as an image-based feature extractor followed by a MLP in end-to-end optimization. 
Zhu $et$ $al$. retrieve prior knowledge shared among distortions, and a deep net is fine-tuned for quality scoring \cite{zhu2020metaiqa}.
Huang $et$ $al$. investigate the inherent relationship between the attributes and the categories via graph convolution network for attribute reasoning and quality estimation \cite{huang2022explainable}. 
Zhang $et$ $al$. optimize a CNN on multiple databases by a hinge constraint on learning uncertainty \cite{zhang2021uncertainty}. 
Li and Huo consider multi-scale visual features and introduce the feedback mechanism \cite{li2022reqa}. 
Chen $et$ $al$. develop multi-scale spatial pooling and combine both block attention and image understanding for improved generalization \cite{chen2022cspp}. 
Sun $et$ $al$. extract low-level features and high-level semantics, and a staircase structure is designed for hierarchical feature integration and quality-aware embedding \cite{sun2021blind}. 
Zhao $et$ $al$. enable representation learning via a pre-text self-supervised task and use a quality-aware contrastive loss to learn distortions \cite{zhao2023quality}. 
Zhang $et$ $al$. design multi-task learning for quality assessment, scene classification and distortion identification \cite{zhang2023blind}. 
Wu $et$ $al$. fuse multi-stage semantic features for no-reference image quality prediction, and before score rating, the features are rectified using multi-level channel attention \cite{wu2024feature}. 

Inspired by the Kolmogorov-Arnold theorem (KAT), a novel module called Kolmogorov-Arnold Networks (KAN) has been proposed \cite{liu2024kan}. 
Except for the activation functions as MLP on nodes, KAN appends learnable activation functions on edges between the nodes of successive layers. 
Improved capacity has been shown in data fitting and knowledge representation \cite{liu2024kanx}. %
Later, KAN variants are designed using different mathematical functions as the activation functions on edges \cite{hou2024comprehensive}.

Despite remarkable success in knowledge representation and data fitting, little is known about KANs for score prediction in BISA. 
This study attempts to bridge this gap. 
Firstly, the Taylor series-based KAN (TaylorKAN) is introduced. 
Then, mid-level features and high-level features of four realistic databases are prepared. 
After that, MLP, SVR and 6 KANs are evaluated. 
Experimental results indicate that KANs are generally better than SVR and MLP, 
and TaylerKAN is the best when using mid-level features as image quality representation.

\section{KAT-inspired networks}
KAT asserts that 
a continuous multivariate function can be represented as a finite sum of continuous uni-variate functions \cite{schmidt2021kolmogorov}. 
Assuming $ f: [0, 1]^n \to \mathbb{R} $ be a continuous function, there exist continuous uni-variate functions \( \phi_{q,p} \) and \( \Phi_q \) that, 
\begin{equation}
f(x_1, x_2, \ldots, x_n) = \sum_{q=1}^{2n+1} \Phi_q \left( \sum_{p=1}^n \phi_{q,p}(x_p) \right).   
\label{kan}
\end{equation}

\subsection{The first KAN model}
KAT-inspired KAN treats a multivariate function as learnable uni-variate spline functions on edges \cite{liu2024kan}. 
A KAN layer can be defined by a matrix of uni-variate functions $ \Phi = \{\phi_{q,p}\} $ in which $ p = 1, 2, \ldots, n_{\text{in}}$ and $ q = 1, 2, \ldots, n_{\text{out}} $. 
Here, $ n_{\text{in}} $ is the input dimension, $ n_{\text{out}} $ is the output dimension, and $ \phi_{q,p} $ is a learnable function. 
The activation of each node in layer $l+1$ can be computed as 
\begin{equation}
x_{l+1,j} = \sum_{i=1}^{n_l} \phi_{l,j,i}(x_{l,i}). 
\label{kanx}
\end{equation}
And thus, the $L$-layered KAN can be generally described as    
\begin{equation}
\text{KAN}(x) = (\Phi_{L-1} \circ \Phi_{L-2} \circ \ldots \circ \Phi_0)(x),  
\label{kany}
\end{equation}
and implemented in a layer-to-layer connection form \cite{liu2024kanx}.

\subsection{KAN variants}
The flexibility of KAN structure allows for diverse implementations using different activation functions on edges. 
The TaylorKAN and other involved KANs are described.  

\subsubsection{TaylorKAN}
Taylor series (Eq. \ref{Taylor}) represent $ f(x) $ as an infinite sum of terms from the values of its derivatives $ f^{(n)} $ at a point $ a $. 
In TaylorKAN, the coefficients are learned during model training. 
\begin{equation}
f(x) \approx \sum_{n=0}^{\infty} \frac{f^{(n)}(a)}{n!} (x-a)^n
\label{Taylor}
\end{equation}

\subsubsection{BSRBF-KAN}
BSRBF-KAN combines B-splines (BSs) and radial basis functions (RBFs) to enhance approximation capabilities \cite{ta2024bsrbf}. 
BSs ensure the continuity, and RBF provides interpolation. 
BSRBF can be represented as, 
\begin{equation}
\phi(x) = w_b b(x) + w_s (\phi_{BS}(x) + \phi_{RBF}(x)), 
\label{BSRBF}
\end{equation} 
where \(w_b\) and \(w_s\) are the weights of BS and RBF respectively, \(\phi_{BS}(x)\) stands for the BS function, and \(\phi_{RBF}(x)\) denotes the RBF function. 
$ \phi_{RBF}(r) = e^{-\epsilon r^2} $ is typically chosen in which \(r\) is the Euclidean distance between the input and the center vector, and \(\epsilon\) controls the width of the Gaussian function.

\subsubsection{ChebyKAN}
ChebyKAN employs the Chebyshev polynomials for function approximation \cite{ss2024chebyshev}. 
It attempts to minimize the maximum error in polynomial approximation for high accuracy and stability. 
Eq. \ref{Cheby} shows Chebyshev polynomials defined by the recurrence relation. 
\begin{equation}
\left\{
\begin{aligned}
& T_0(x) = 1 \\
& T_1(x) = x \\
& T_{n+1}(x) = 2xT_n(x) - T_{n-1}(x) 
\end{aligned}
\right.
\label{Cheby}
\end{equation}

\subsubsection{HermiteKAN}
Hermite polynomials are ideal for approximating these Gaussian-like functions due to their recurrence relations and orthogonality. 
HermiteKAN is based on Hermite polynomials (Eq. \ref{Hermite}).  
\begin{equation}
H_n(x) = (-1)^n e^{x^2} \frac{d^n}{dx^n} e^{-x^2} 
\label{Hermite}
\end{equation}

\subsubsection{JacobiKAN}
JacobiKAN uses Jacobi polynomials \cite{aghaei2024fkan}. 
The polynomials are orthogonal to the weight function and flexible in handling diverse boundary conditions. 
\begin{equation}
P_n^{(\alpha,\beta)}(x) = \frac{1}{2^n} \sum_{k=0}^n \binom{n+\alpha}{k} \binom{n+\beta}{n-k} (x-1)^{n-k} (x+1)^k 
\label{Jacobi}
\end{equation}

\subsubsection{WavKAN}
Wavelets are adept at capturing local variations in functions, and they provide flexibility and adaptability in complex pattern modeling. 
The general wavelet transformation form is shown as below, 
\begin{equation}
\psi_{a,b}(x) = \frac{1}{\sqrt{a}} \psi \left( \frac{x-b}{a} \right),
\label{wave}
\end{equation} 
where \(a\) and \(b\) correspond to the scaling and the translation parameters. 
Since different wavelet basis are available, 
in this study, the Mexican Hat wavelets are used in WavKAN \cite{bozorgasl2024wav}, 
\begin{equation}
\psi(x) = \frac{2}{\sqrt{3\pi^{1/4}}} \left( 1 - x^2 \right) e^{-\frac{x^2}{2}}.
\label{Mexican}
\end{equation}

\subsubsection{Other variants}
Other KANs are accessible \cite{hou2024comprehensive}. 
However, limited by computing resources, above-mentioned light-weight KANs are evaluated in the current study.  

\section{Materials and Methods}

\subsection{Databases}
Four databases (BID2011 \cite{ciancio2010no}, CID2013 \cite{virtanen2014cid2013}, CLIVE \cite{ghadiyaram2015massive} and KonIQ-10k \cite{hosu2020koniq}) with realistic distortions are analyzed. 
Table \ref{datasets} shows the number (\#) of distorted images, the year of data availability, and the score ranges of the databases. 
\begin{table}[h]  %
  \centering
  \caption{General information of the databases}
    \begin{tabular}{llll}
	\hline
          			                                   & \# images	     & year  & score range 	\\
	\hline
    	BID2011    \cite{ciancio2010no}	              & 586   	        & 2011  & [0, 5] 		\\
    	CID2013	   \cite{virtanen2014cid2013}	      & 474   	        & 2013  & [0, 100] 		\\
    	CLIVE      \cite{ghadiyaram2015massive}       & 1,169           & 2015  & [0, 100] 	\\
    	KonIQ-10k \cite{hosu2020koniq}	              & 10,073          & 2018  & [0, 5] 		\\
    \hline
    \end{tabular}%
  \label{datasets}%
\end{table}%

\subsection{Feature preparation}
Two sets of features are prepared. One contains 15 mid-level features per image that are the output of BISA indicators \cite{yu2023hybrid}.
The other set includes 2048 deeply learned features which are derived from the last full-connection layer of the pre-trained ResNet50 \cite{he2016deep}. 
The procedure is similar as \cite{wu2024feature}. 

\subsubsection{Mid-level features}
Assuming a database $\{ (I_i, y_i) \}_{i=1}^n$ with $n$ pairs of images and scores, an indicator $\eta_j$ yields an objective score $x_{i,j}$ to an image $I_i$, which can be formulated as $x_{i,j} = \eta_j(I_i)$. 
In the same way, $m$ indicators ($\{\eta_j\}_{j=1}^m$) generate a feature matrix $M$ as shown in Eq. \ref{matrixM}. 
\begin{equation}
M = 	
\left[ 
		\begin{array}{ccc|c}
			x_{1,1}		& \dots  	& x_{1,m} 		& y_1		\\
		 	\vdots  	& \ddots  	& \vdots 		& \vdots	\\
			x_{n,1} 	& \dots		& x_{n, m}		& y_n
		\end{array}
\right]_{n \times (m+1)}
\label{matrixM}
\end{equation}

\subsubsection{High-level features}
Using pre-trained ResNet50 \cite{he2016deep} as an extractor $f$, it generates a $d$-dimensional feature vector ($ d = 2048 $) to an input image $I_i$. 
Thereby, deeply learned high-level feature matrix $N$ is derived as shown in Eq. \ref{matrixN}.
\begin{equation} 
N = 	
\left[ 
		\begin{array}{c|c}

			f(I_1) 			& y_1		\\
		 	\vdots  	 		& \vdots	\\
			f(I_i)				& y_i 		\\
			\vdots  	 		& \vdots	\\
			f(I_n) 			& y_n
		\end{array}
\right]_{n \times (d+1)}
\label{matrixN}
\end{equation}

\subsection{Score prediction}
Besides KANs, SVR and MLP are tested. 
Assume features ($ X = \{ \vec{x_i} \}_{i=1}^{l}$) 
and scores ($ Y = \{ y_i \}_{i=1}^{l}$) of $l$ samples in the training set, 
a weighting vector $ \boldsymbol{w} $, 
and a new input vector $\vec{x}$. 
SVR ($svr$) aims to find a function $R_{svr}(X)$ that maximizes the deviation of $\epsilon$ 
from the subjective score $y_i$ of training samples. 
In Eq. \ref{svr}, $\zeta(X)$ is a nonlinear function, and $\gamma$ is a bias. 
\begin{equation}
g(X) = R_{svr}(X) = \boldsymbol{w}^T \zeta(X) + \gamma
\label{svr}
\end{equation}

MLP ($mlp$) is designed with different numbers of the hidden layers in terms of different feature inputs. 
Its parameters are optimized by minimizing the difference between the output of model $R_{mlp}$ and the ground truth $Y$. 
\begin{equation}
	\boldsymbol{w}^* = \text{arg min} \quad || Y - R_{mlp}(\boldsymbol{w}; X) ||^2
	\label{mlp}
\end{equation}

\subsection{Performance criterion}
The performance is evaluated by using Pearson linear correlation coefficient (PLCC) and Spearman rank order correlation coefficient (SRCC). 
Higher values indicate better performance. 

Specifically, PLCC is computed after a five-parameter nonlinear mapping between objective and subjective scores (Eq. \ref{fitting}) 
in which $s$ denotes the predicted score, $f(s)$ is the mapped score, 
and $ \{q_i\}_{i=1}^{5} $ are the fitting parameters. 
PLCC values are calculated between subjective scores and mapped scores. 
\begin{equation}
f(s) = q_1 (\frac{1}{2} - \frac{1}{1+ e^{q_2(s-q_3)}}) +q_4 s + q_5 
\label{fitting}
\end{equation}

Training time is used to evaluate the efficiency of involved score prediction models. 
The training time for SVR is recorded in seconds ($s$), 
and the other models are measured in iterations per second ($i/s$), 
where a lower value indicates poorer computational efficiency.

\subsection{Implementation details} 
In each experiment, a database is randomly split into three subsets for training (70\% samples), validation (15\% samples) and testing (the remaining samples) of BIQA indicators. 

For fair comparison, the KANs and MLP are configured with 3 hidden layers ([15, 26, 18, 12, 1]) for mid-level inputs and 4 hidden layers ([2048, 1536, 1024, 256, 128, 1]) for high-level feature inputs. TaylorKAN is implemented with quadratic approximation, and the other KAN models use default settings. To SVR, the radial basis function (RBF) kernel is used, and the other parameters are set with default values. 


During training, an early stopping mechanism is used even though 500 iterations are pre-defined. 
The patience parameter is set to 20, and thus, 
the training will be terminated early if the validation loss fails to improve for 20 consecutive epochs. 
Throughout the process, the metrics to the learning rate that yields the highest sum of PLCC and SRCC are reported.

The codes are implemented on an Ubuntu operating system (version 22.04) using Python (version 3.12), PyTorch (version 2.3.0), and CUDA (version 12.1). The algorithms are executed on a GPU (A100-PCIE-40GB) with 72 GB RAM. The project is available at https://github.com/CUC-Chen/KAN4IQA.

\section{Results and Discussion}
Experimental results using mid-level features and high-level features are respectively shown in Table \ref{mid} and \ref{deep} where the $^{\dagger}$ denotes the training time of SVR in $s$. 
Table \ref{real} compares the current study with several state-of-the-art achievement. 
In the tables, the best metric values are boldfaced. 

\subsection{BISA using mid-level features}
Table \ref{mid} shows that compared to SVR, 
TaylorKAN achieves better prediction on BID2011, CID2013 and KonIQ-10k, and its PLCC value is higher on CLIVE. The other KAN models obtain better performance on BID2011, competitive results on CID2013 and KonIQ-10k, but worse prediction on CLIVE. Among the KANs, ChebyKAN is the fastest, achieving over 30 $i/s$ on BID2011 and CID2013, and TaylorKAN demonstrates slightly lower but comparable computational efficiency.
\begin{table*}[htbp]
  \centering
  \caption{BISA by using 15 mil-level features}
    \begin{tabular}{l|ccc|ccc|ccc|ccc}
\hline
    & \multicolumn{3}{c|}{BID2011 \cite{ciancio2010no} } 					
    & \multicolumn{3}{c|}{CID2013 \cite{virtanen2014cid2013} } 
    & \multicolumn{3}{c|}{CLIVE \cite{ghadiyaram2015massive} } 		
    & \multicolumn{3}{c} {KonIQ-10k \cite{hosu2020koniq} }                            \\  
        & PLCC                      & SRCC                  & time ($i/s$)   				
        & PLCC                      & SRCC                  & time ($i/s$)			
        & PLCC                      & SRCC                  & time ($i/s$)  				
        & PLCC                      & SRCC                  & time ($i/s$)          \\
\hline
    SVR  									
        & 0.619                     & 0.617                 & 0.049$^{\dagger}$          	
        & 0.834                     & 0.810                 & {0.024}$^{\dagger}$  
        & 0.630                     & \textbf{0.592}        & {0.117}$^{\dagger}$         		
        & 0.746                     & 0.691 	            & {12.39}$^{\dagger}$              \\
    MLP  									
        & 0.744                     & 0.729                 & 33.66      		
        & 0.808                     & 0.791                 & \textbf{39.62}       	
        & 0.649                     & 0.552                 & 18.06       		
        & 0.753                     & 0.682                 & 2.082                 \\ 		
\hline
    BSRBF\_KAN \cite{ta2024bsrbf} 								
        & 0.675                     & 0.680                 & 33.23        				
        & 0.845                     & 0.795                 & 12.99        						
        & {0.562}                   & 0.479                 & 6.347        	
        & 0.725                     & 0.650                 & 1.463                 \\  	
    ChebyKAN \cite{ss2024chebyshev}								
        & 0.700                     & 0.703                 & \textbf{34.86}        	
        & 0.808                     & 0.826                 & {37.73}        						
        & 0.570                     & 0.447                 & \textbf{18.36}   
        & 0.749                     & 0.680                 & \textbf{2.224}        \\ 			
    HermiteKAN 								
        & {0.651}                   & {0.740}               & 19.29            
        & {0.825}                   & {0.845}               & 20.99  
        & 0.566                     & 0.502                 & 9.655       	 			
        & 0.754                     & {0.671}               & 1.118                 \\ 
    JacobiKAN \cite{aghaei2024fkan}								
        & 0.709                     & \textbf{0.789}        & 16.29        	
        & 0.808                     & {0.775}               & 20.12        			
        & 0.545                     & {0.519}               & 9.546        	
        & {0.753}                   & 0.689                 & 1.074                 \\  	 
    WavKAN \cite{bozorgasl2024wav} 									
        & 0.715                     & 0.730                 & 22.74  
        & 0.827                     & 0.827                 & 26.60       						
        & 0.559                     & 0.482                 & 13.23        	
        & {0.759}                   & {0.685}               & 1.448                 \\  		
    TaylorKAN (ours)								
        & \textbf{0.756}            & 0.782                 & 28.44	        	
        & \textbf{0.871}            & \textbf{0.851}        & 34.45  
        & \textbf{0.668}            & 0.582                 & 15.32        	
        & \textbf{0.766}            & \textbf{0.699}        & 1.927               \\  %
\hline
    \end{tabular}%
  \label{mid}%
\end{table*}%


When using the 15 mid-level features for quality representation, TaylorKAN achieves the best performance, attributed to several factors. Firstly, it surpasses SVR by leveraging the Taylor series for hierarchical approximation, and meanwhile, a multi-layer network is implemented for deeper feature learning. Secondly, it outperforms other involved KANs, possibly because quadratic approximation enables more precise feature fitting. Thirdly, it achieves higher metric values than MLP, since the former uses additional activation functions on edges. At last, the simplicity of the quadratic approximation enables TaylorKAN to perform data fitting at a relatively faster pace.

\subsection{BISA using high-level features}
Setting SVR as the baseline, Table \ref{deep} suggests that 
KANs achieve higher values on BID2011, worse results on CID2013 and CLIVE, 
and close performance on KonIQ-10k in general. 
Notably, BSRBF\_KAN and WavKAN are comparable to SVR on CID2013 and CLIVE, 
while TaylorKAN, BSRBF\_KAN, JacobiKAN and HermiteKAN are superior on KonIQ-10k. 
It is found that BSRBF\_KAN, WavKAN and TaylorKAN run at less than 10 $i/s$ on BID2011 and CID2013. 
\begin{table*}[htbp]
  \centering
  \caption{BISA by using 2048 high-level features}
    \begin{tabular}{l|ccc|ccc|ccc|ccc}
\hline
    & \multicolumn{3}{c|}{BID2011 \cite{ciancio2010no} } 					
    & \multicolumn{3}{c|}{CID2013 \cite{virtanen2014cid2013} } 
    & \multicolumn{3}{c|}{CLIVE \cite{ghadiyaram2015massive} } 		
    & \multicolumn{3}{c} {KonIQ-10k \cite{hosu2020koniq} }                            \\  
          & PLCC                & SRCC                      & time ($i/s$)   				
          & PLCC                & SRCC                      & time ($i/s$)   			
          & PLCC                & SRCC                      & time ($i/s$)   								
          & PLCC                & SRCC                      & time ($i/s$) \\
\hline
    SVR
          & 0.786               & 0.782                     & {0.355}$^{\dagger}$         
          & \textbf{0.860}      & \textbf{0.882}            & {0.447}$^{\dagger}$        
          & {0.751}             & \textbf{0.712}            & 1.492$^{\dagger}$         
          & 0.839               & 0.800                     & 119.22$^{\dagger}$                 \\ 		
    MLP  									
          & 0.750               & 0.780                     & \textbf{23.46}         
          & 0.796               & 0.825                     & 18.32        
          & 0.637               & 0.554                     & \textbf{12.77}         
          & 0.808               & 0.763                     & 1.038                      \\ 
\hline
    BSRBF\_KAN \cite{ta2024bsrbf} 
          & 0.811               & 0.795                     & 6.296        
          & {0.828}             & {0.820}                   & 6.247        
          & 0.733               & 0.649                     & 3.768       
          & 0.841               & {0.809}                   & 0.435                      \\ 
    ChebyKAN \cite{ss2024chebyshev}
          & {0.821}             & 0.812                     & 20.79        
          & 0.630               & 0.665                     & \textbf{18.61}        
          & 0.662               & 0.587                     & 11.41        
          & 0.824               & 0.790                     & \textbf{1.368}         \\ 		
    HermiteKAN 
          & \textbf{0.822}      & \textbf{0.814}            & 12.53       
          & 0.604               & 0.687                     & 10.45        
          & 0.670               & 0.647                     & 6.804        
          & {0.839}             & 0.804                     & 0.803                        \\  	
    JacobiKAN \cite{aghaei2024fkan}
          & {0.820}             & {0.806}                   & 11.38        
          & 0.596               & 0.605                     & 10.65        
          & {0.733}             & {0.651}                   & 6.426          
          & {0.842}             & {0.803}                   & 0.699                     \\  
    WavKAN \cite{bozorgasl2024wav}
          & 0.767               & 0.735                     & 1.178        
          & {0.844}             & {0.856}                   & 0.994      
          & \textbf{0.752}      & 0.676                     & 0.587       
          & {0.810}             & {0.777}                   & 0.067                     \\  
    TaylorKAN (ours)
          & 0.797               & {0.813}                   & 3.285        
          & 0.788               & 0.780                     & 2.796        
          & 0.696               & 0.598                     & 1.644        
          & \textbf{0.850}      & \textbf{0.811}            & 0.188 \\ 
\hline
    \end{tabular}%
  \label{deep}%
\end{table*}%


When applying the 2048 high-level features for image quality representation, TaylorKAN obtains slightly higher metric values on BID2011 and KonIQ-10k, while worse results on CID2013 and CLIVE than SVR. The sub-optimal performance indicates that it remains challenging for the KANs to handle high-dimensional features. On the one hand, the KAT (Eq. \ref{kan}) implies that the representation of a high-dimensional function requires a large number of uni-variate functions. Consequently, the accumulation of errors and over-fitting may cause unstable data-driven approximation. On the other hand, whether these high-level features primarily learned for object recognition can be directly utilized to represent image quality has not yet been determined. Therefore, proper post-processing strategies, such as fine-tuning \cite{zhu2020metaiqa} and rectification \cite{wu2024feature}, 
become important for smoothly transferring these deeply learned features into the representation space of image quality.

Comparison of Table \ref{mid} and Table \ref{deep} reveals that the 2048 high-level features provide more effective quality representation than the 15 mid-level features for score prediction, except on the CID2013 database. Most KANs, such as BSRBF\_KAN, ChebyKAN and JacobiKAN, achieve better performance when high-level features are used, while the utilization of the 2048 features dramatically increases the computational time. On the one hand, it should be acknowledged the mid-level features are less effective than the high-level features due to their limited quantity and capacity for quality representation. Specifically, the mid-level features are primarily designed for scoring image sharpness \cite{yu2023hybrid}, while the high-level features are hierarchically and deeply learned for general recognition task \cite{he2016deep}. On the other hand, the KANs perform generally better on CID2013 when using the mid-level features. For example, in the case of JacobiKAN, using mid-level features leads to PLCC 0.808 and SRCC 0.775, whereas using high-level features yields PLCC 0.596 and SRCC 0.605. This finding indicates that the mid-level features are more suitable for the CID2013 database.

\subsection{Current achievement on the databases}
The results of TaylorKANs using mid-level features ($^a$) and high-level features ($^b$) and several state-of-the-art (SOTA) works are shown in Table \ref{real}. These works \cite{zhang2021uncertainty, huang2022explainable, li2022reqa, chen2022cspp, sun2021blind, zhao2023quality, zhang2023blind} develop novel CNNs that take images rather than features as input. 
\begin{table*}[htbp]
  \centering
  \caption{State-of-the-art performance on realistic distorted image datasets}
    \begin{tabular}{l|cc|cc|cc|cc}
\hline
        & \multicolumn{2}{c|}{BID2011 \cite{ciancio2010no} } 							
        & \multicolumn{2}{c|}{CID2013 \cite{virtanen2014cid2013} }
        & \multicolumn{2}{c|}{CLIVE \cite{ghadiyaram2015massive} } 				
        & \multicolumn{2}{c} {KonIQ-10k \cite{hosu2020koniq} } 	          \\
\hline
        & PLCC  & SRCC  	
        & PLCC  & SRCC   		
        & PLCC  & SRCC  		
        & PLCC  & SRCC                         \\
\hline          
        TaylorKAN $^{a}$ 								
                & 0.756                     & 0.782                         	
                & \textbf{0.871}            & \textbf{0.851}         
                & {0.668}                   & 0.582                         	
                & {0.766}                   & {0.699}                                  \\   
        TaylorKAN $^{b}$ 
                & \textbf{0.797}            & \textbf{0.813}          
                & 0.788                     & 0.780                 
                & \textbf{0.696}            & \textbf{0.598}              
                & \textbf{0.850}            & \textbf{0.811}                 \\ 
\hline
	  SARQUE \cite{huang2022explainable} 	
                & 0.861                     & 0.846  
                & \textbf{0.934}            & \textbf{0.930} 
                & 0.873                     & 0.855     
                & 0.923                     & 0.901                                         \\ 
	  UNIQUE \cite{zhang2021uncertainty}		
                & 0.873                     & 0.858 
                &                           & 
                & 0.890                     & 0.854     
                & 0.901                     & 0.896                                          \\ 	                
	  REQA \cite{li2022reqa}  				
                & 0.886                     & 0.874  
                &                           & 
                & 0.880                     & 0.865      
                & 0.916                     &  0.904                                          \\
	  CSPP-IQA \cite{chen2022cspp} 			
                & {0.891}                   & {0.875}   
                &                           &  
                & 0.898                     & 0.882     
                & 0.921                     & 0.912                                          \\        
	  StairIQA \cite{sun2021blind} 			
                & \textbf{0.9284}           & \textbf{0.9128}     
                &                           &  
                & \textbf{0.9175}           & {0.8992}        
                & {0.9362}                  & {0.9209}                      \\
        QPT-ResNet50 \cite{zhao2023quality} 			
                & {0.9109}                  & {0.8875}     
                &                           &  
                & {0.9141}                  & {0.8947}        
                & \textbf{0.9413}           & \textbf{0.9271}                     \\  
        LIQE \cite{zhang2023blind} 			
                & {0.900}                   & {0.875}    
                &                           & 
                & {0.910}                   & \textbf{0.904}        
                & {0.908}                   & {0.919}                     \\  
\hline 
    \end{tabular}%
  \label{real}%
\end{table*}%

A significant performance gap is found between the SOTA works and TaylorKAN in score prediction. The CNNs achieve PLCC $\ge$ 0.86 and SRCC $\ge$ 0.84, which are much higher than those of the evaluated models (Table \ref{mid} and \ref{deep}). These SOTA works benefit not only from advanced architectures of image convolution and feature pooling but also from sophisticated module designs, such as uncertainty learning \cite{zhang2021uncertainty}, spatial pooling \cite{chen2022cspp}, self-supervised learning \cite{zhao2023quality}, and multitask learning \cite{zhang2023blind}. Thus, one way to improve representation power and score prediction performance is to integrate prior knowledge and advanced modules into KANs \cite{hou2024comprehensive}. When using TaylorKAN for score prediction, high-level features can increase, but not substantially, metric values on three databases. On the one hand, the perceptual similarity is an emergent property shared across deep visual embedding \cite{zhang2018unreasonable} and thus, it is not surprise that deep features are effective for image quality representation \cite{zhao2023quality}. On the other hand, how to figure out the most informative subsets of deep features is important for KANs to enhance effectiveness and efficiency in score prediction, while avoiding the challenges of high-dimensional feature processing. 

This study has several limitations. Firstly, a broader range of informative features, potentially numbering in the hundreds or thousands, could be collected and refined by selecting the most relevant ones. These selected features can then be applied in KANs for further data fitting and feature embedding tailored to specific tasks. Secondly, KANs could be integrated into deep learning architectures by replacing the MLP components. Such integration may enhance KANs' representation capacity in end-to-end optimization. Thirdly, the performance of KANs could be further assessed on other tasks, such as object recognition, to gain deeper insights into their strengths, limitations, and potential applications in feature representation.

\section{Conclusions}
In addition to SVR and MLP, six KANs are evaluated on four realistic databases for score prediction respectively using mid-level and high-level features. The results demonstrate that KANs achieve superior or competitive performance compared to SVR and MLP, highlighting their potential for enhancing performance in score prediction and related tasks.

\end{document}